\documentclass[letterpaper,twocolumn,fleqn]{article} 

\usepackage{ist}
\usepackage{amsfonts, amsmath}
\usepackage[mathscr]{euscript}
\usepackage{color}
\title{GAN Based Image Deblurring Using Dark Channel Prior}

\author{ Shuang Zhang$^*$, Ada Zhen, Robert L. Stevenson; University of Notre Dame; Notre Dame, Indiana, 46556}

\date{} 

\begin{document} 

\maketitle 

\thispagestyle{empty} 


\begin{abstract}
	A conditional general adversarial network (GAN) is proposed for image deblurring problem. It is tailored for image deblurring instead of just applying GAN on the deblurring problem. Motivated by that, dark channel prior is carefully picked to be incorporated into the loss function for network training. To make it more compatible with neuron networks, its original indifferentiable form is discarded and L2 norm is adopted instead. On both synthetic datasets and noisy natural images, the proposed network shows improved deblurring performance and robustness to image noise qualitatively and quantitatively. Additionally, compared to the existing end-to-end deblurring networks, our network structure is light-weight, which ensures less training and testing time.
	
\end{abstract}

\section{Introduction}
\label{sec:intro}

Blur is a common artifact for images taken by hand-held cameras. It is mostly caused by object motions, hand shake or out-of-focus. The blurry image is often modeled as convolution of a sharp image and a blur kernel. And the target of deblurring is to restore a latent sharp image from the blurry one. Single image deblurring, however, is a highly ill-posed problem, since it contains insufficient information to recover a unique sharp image.

In the past few years, assorted constraints and regulation schemes have been proposed to exclude implausible solutions. Priors, like total variation prior \cite{PriorTV}, sparse image prior \cite{PriorSI}, heavy-tailed gradient prior \cite{PriorHT} and dark channel prior \cite{PriorDC}, are combined with $L_1/L_2$ norm image regulation term to suppress ringing artifacts and improve quality. Zhen \cite{Ruiwen} takes advantage of inertia sensor data to gain extra information and estimate spatially varying blur kernels. However, since the blur kernel in reality is more complicated than the model, estimation of blur kernel is inaccurate, which causes ringing artifacts. Furthermore, these methods based on iterative optimization techniques are computationally intensive.

Recently, Convolutional Neural Networks (CNN) and deep learning related techniques have drawn a great attention in computer vision and image processing. Their applications in image deblurring demonstrate promising results. Sun \cite{Sun} and Schuler \cite{Schuler} use CNN to estimate the spatially-invariant blur kernel and obtain latent image by tradition pipeline. Chakrabarti \cite{Chakra} trained a neural network to predict complex Fourier coefficients of motion kernel. Recently kernel-free end-to-end deblurring methods are proposed by Nah et al. \cite{Nah} and Kupyn et al. \cite{Kupyn}. Nah \cite{Nah} adopted the multi-scale network to mimic conventional coarse-to-fine optimization methods, and proposed a new realistic blurry image dataset with ground truth sharp images. The work of Kupyn \cite{Kupyn} trains the popular Generative Adversarial Network (GAN) on the same dataset with fewer parameters, gains higher PSNR values than that of Nah et al. \cite{Nah} on the GOPRO dataset, and beats the others on Kohler dataset \cite{Kohler} using SSIM. Although \cite{Kupyn} performs well based on metric scores, visually, its deblurred result suffers grid artifacts, as illustrated in Fig. 1.

\begin{figure}[!t]
	\includegraphics[width=\columnwidth]{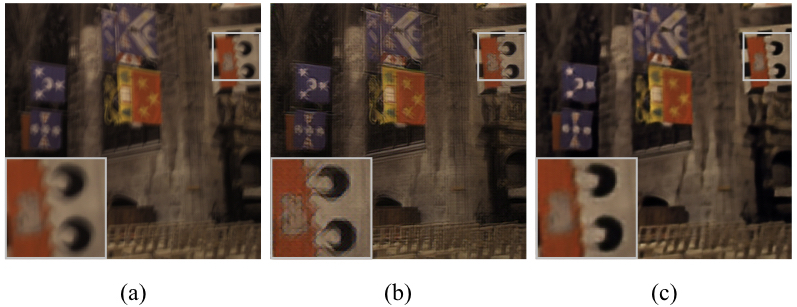}
	\caption{Comparison. (a) Input blurry image. (b) Result of \cite{Kupyn}. (c) Our result.}
	\label{Fig:01}
\end{figure}

To address this artifact, we utilize the dark channel prior. Dark channel is defined as minimal intensity among three color channels of pixels in a local area. It was first proposed by He et al. \cite{He} for dehazing problem, based on the statistics that haze-free outdoor images have a smaller dark channel than hazy images. Pan et al. \cite{PriorDC} applied dark channel prior to image deblurring. They theoretically and empirically proved that comparing with blur images, the dark channel of sharp image is more sparse. And their results demonstrate that dark channel prior contributes to suppressing ringing and other artifacts. In order to enforce the sparsity, they utilize a regulation term of $L_0$ norm to count the nonzero elements of dark channel maps. Unfortunately, $L_0$ norm is not differentiable, which makes it hard to utilize in back propagation of neuron networks. Instead of using $L_0$ norm, we adopt $L_2$ norm to directly compute difference of the dark channel maps between groundtruth sharp images and deblurred images.

In this paper, we present a GAN based image deblurring network using dark channel difference as loss function. The proposed technique is not just a straightforward application of GAN. This method focuses on how to combine traditional knowledge with deep learning to make the network achieve better performance. Compared to the previous GAN-based deblurring network, the proposed network has less layers and weights. It leads to less training and testing time, and more importantly achieves favorable results. In addition, the original GOPRO training dataset consists of artificially created blurry images without noise, which are usually different from the real blurry images. To improve the quality of our trained network on more realistic blurry images and increase network robustness, we add random Gaussian noise with variance in a limited range onto the training image patches. The comparison experiments show that our network outperforms Kupyn et al. \cite{Kupyn} for both GOPRO test dataset and real noisy blurry images.

\begin{figure*}[!t]
	\includegraphics[width=2\columnwidth]{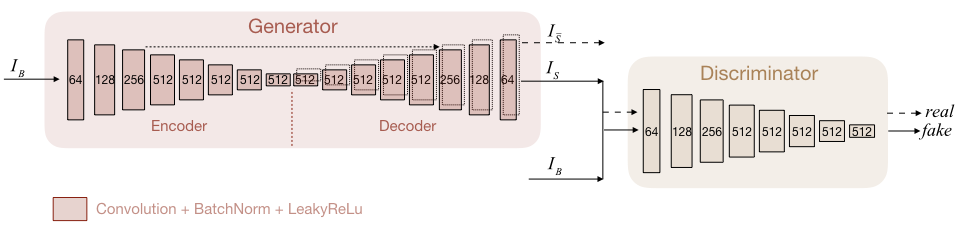}
	\caption{Proposed Network. The proposed CGAN based network has two sub-networks: generator $G$ and discriminator $D$. Generator restores sharp image $I_S$ from input blurry image $I_B$. $I_{\bar{S}}$ represents ground truth image. Discriminator can regard input pair $(I_S, I_B)$ as "fake" and $(I_{\bar{S}}, I_B)$ as "real". Except for the first layer of discriminator and generator, each block in both generator and discriminator consists of a convolutional layer, batch normalization step \cite{BatchNorm} and an activation function LeakyReLu \cite{LReLu}. The first layers are not normalized. The digit denotes the number of filters for each block. Dotted lines are skip connection layers in decoder which come from layers with same size in encoder.}
	\label{Fig:02}
\end{figure*}


\section{Related Work}
\subsection{Conditional General Adversarial Networks}

GAN is first proposed by Goodfellow et al. \cite{Goodfellow} to train a generative network in an adversarial process. It consists of two networks: a generator $G$ and a discriminator $D$.  Generator generates a fake sample from input noise $z$, while discriminator estimates the probability that the fake sample is from training data rather than generated by generator. These two networks are simultaneously trained until discriminator cannot tell if the sample is real or fake. This process can be summarized as a two-player min-max game with the following function:
\begin{equation}
\min_G \max_D \mathbb{E}_{\bar{x}\sim \text{P}_{\text{data}}(\bar{x})} \left[ \log D(\bar{x})\right] + 
\mathbb{E}_{z\sim \text{P}_{z}(z)} \left[ \log (1-D(G(z))) \right],
\end{equation}
where $\text{P}_{\text{data}}$ denotes distribution over training data $\bar{x}$ and $\text{P}_{z}$ is distribution of input noise $z$. GAN has been applied to different image restoration problems like super-resolution \cite{Superresolution} and texture transfer \cite{Styletrans}. 

Mirza et al. \cite{CGAN} extend GAN into a conditional model (eq. (\ref{eq:CGAN})), called Conditional Generative Adversarial Nets (CGAN), so that GAN can make use of auxiliary information to direct both generator and discriminator. Isola et at. \cite{Pix2pix} adopt CGAN architecture to achieve general image-to-image translation. In \cite{Pix2pix}, more than just random noise $z$, similar image $y$ is added as input of the generator, where $y$ and $\bar{x}$ share part of features. $y$ and $\bar{x}$ can be pairs of hazing and clear images about same scene, or different color buildings with same structure. Based on network architecture of \cite{Pix2pix}, Kupyn et al. \cite{Kupyn} utilize Wasserstein loss \cite{WGAN} and perceptual loss \cite{Perceptual} to train a CGAN for deblurring problem.
\begin{equation} \label{eq:CGAN}
\min_G \max_D \mathbb{E}_{\bar{x}\sim \text{P}_{\text{data}}(\bar{x})} \left[ \log D(\bar{x}, y)\right] + 
\mathbb{E}_{z\sim \text{P}_{z}(z)} \left[ \log (1-D(G(z,y),y)) \right],
\end{equation} 
\subsection{Dark Channel Prior}
For an image $I$, the dark channel of a pixel $p$ is defined by He et al. \cite{He} as
\begin{equation} \label{eq:DC}
\mathfrak{D}_c(p) = \min_{q \in \mathcal{N}(p)} \left( \min_{c \in \{r,g,b\}} I^{c} (q)\right),
\end{equation}
where $p$ and $q$ are pixel locations, $\mathcal{N}(p)$ denotes the image patch centered at $p$, and $I^c$ is the $c$-th color channel. As shown in eq. (\ref{eq:DC}), dark channel describes the minimum intensity in an image patch. He et al. \cite{He} observe that dark channel map $D(I)$ in a haze-free image tends to be zero. Pan et al. \cite{PriorDC} use a less restrictive assumption that dark channel map $D(I)$ is sparse rather than zero. Inspired by this, they adopt $L_0$-regulation term to enforce less sparse dark channel in a deblurring process, where $L_0$ norm counts non-zero elements in a dark channel map.

\section{Proposed Method}

\subsection{Network Architecture}
The proposed network aims at obtaining a generator to restore sharp image $I_S$ from input blurry image $I_B$. This generator is trained with a discriminator using pairs of blurry image $I_B$ and ground truth sharp image $I_{\bar{S}}$. This structure is shown in fig.\ref{Fig:02}. Except for the first layer of discriminator and generator, each block in both generator and discriminator consists of a convolutional layer, batch normalization step \cite{BatchNorm} and an activation function LeakyReLu \cite{LReLu} with leaking rate $\alpha = 0.2$. The first layers are not normalized.

{\bf Generator} The proposed generator adopts an encoder-decoder framework to achieve image-to-image performance. Similar to \cite{Pix2pix}, the encoder consists of a sequence of convolutional layers with stride $=2$ and kernel size $=5$. And the decoder has a chain of transposed-convolutional layers with same size of stride and kernel. Encoder represents the input image with a bottleneck vector and decoder recovers an image with same size of input from bottleneck vector. A skip architecture is applied by inserting same size of layers from encoder after each layer of decoder.  This skip connection refines the details in output image by combining deep, coarse, semantic information and shallow, fine, appearance information \cite{FC}. Dropout is also included in decoder to avoid over-fitting.

{\bf Discriminator} The proposed discriminator contains a series of convolutional layers with stride $=2$ and kernel size $=5$. The output of discriminator is a scalar, followed by a sigmoid function.

\begin{figure*}[!t]
	\includegraphics[width=2\columnwidth]{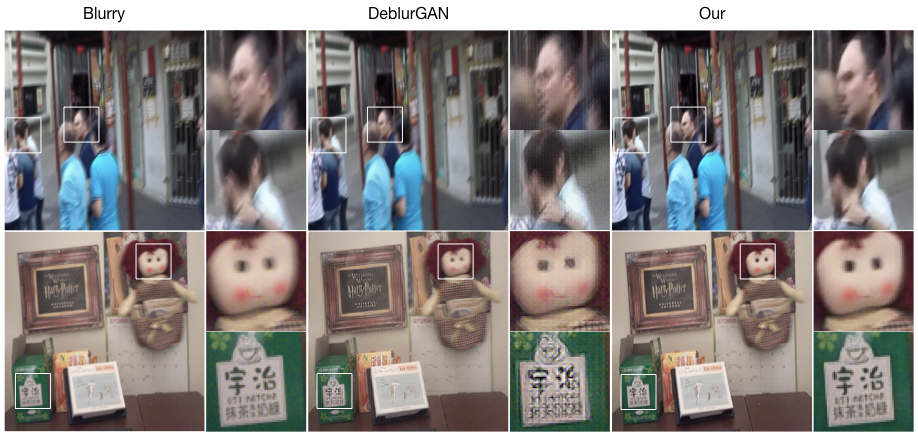}
	\caption{Comparison with DeblurGAN \cite{Kupyn}. From top to bottom: image from GOPRO dataset and real nature image. From left to right: blurry images, deblurred results by \cite{Kupyn} and our result.} 
	\label{Fig:03}
\end{figure*}
\subsection{Loss Functions}
According to eq. (\ref{eq:CGAN}), we train discriminator and generator alternatively. The loss function of discriminator is same as adversarial loss:
\begin{equation} \label{eq:d_loss}
\mathscr{L}_d = \mathbb{E}_{\bar{x}, y} \left[ \log D(\bar{x}, y)\right] + 
\mathbb{E}_{y,z} \left[ \log (1-D(G(z,y),y)) \right].
\end{equation} 
In the deblurring setting, $y$ and $\bar{x}$ denote blurry and sharp image, respectively.
The generator loss is defined as combination of adversarial loss, content loss and dark channel loss:
\begin{equation} \label{eq:g_loss}
\mathscr{L}_g =\mathbb{E}_{y,z} \left[ \log (1-D(G(z,y),y)) \right] + \lambda_1 \cdot \mathscr{L}_{content} + \lambda_2 \cdot \mathscr{L}_{darkchannel},
\end{equation}
where $\lambda_1 = 100$ and $\lambda_2 = 250$ in our experiments. 

{\bf Content loss} We adopt the traditional content loss to direct the output of generator to ground truth. Although both $L_1$ and $L_2$ norm are commonly used, $L_1$ norm is chosen since it attains less blurring result \cite{Pix2pix}.
\begin{equation} \label{eq:L1_loss}
\mathscr{L}_{content} =  \mathbb{E}_{\bar{x}, y, z} \left[ ||\bar{x}-G(y,z) ||_1\right].
\end{equation}

{\bf Dark channel loss} In order to suppress ringing and grid artifact, dark channel prior is especially chosen. Pan et al. \cite{PriorDC} exploit $L_0$ norm to count non-zero elements in a dark channel map  $\mathfrak{D}_c(I)$ of an image $I$. Since $L_0$ norm is indifferential, $L_2$ norm is utilized instead which calculates the distance of dark channel map between ground truth and deblurred image.
\begin{equation} \label{eq:DC_loss}
\mathscr{L}_{darkchannel} =  \mathbb{E}_{\bar{x}, y, z} \left[ ||\mathfrak{D}_c(\bar{x})-\mathfrak{D}_c(G(y,z)) ||_2\right].
\end{equation}

Unlike \cite{Kupyn}, we discard the perceptual loss \cite{Perceptual}. Kupyn et al. \cite{Kupyn}
employ the difference of one feature map in the VGG-19 \cite{Vgg19} between ground truth and restored images as perceptual loss. GAN is known for its ability to reserve perceptual feature of an image. Adding an extra perceptual loss seems a noneffective repeat. Our experiment shows that perceptual loss doesn't improve the result, on the contrary, it leads to worse performance.


\section{Experiments}
Our network is implemented with Python code based on Tensorflow \cite{Tensorflow}. 
\subsection{Datasets}
GOPRO dataset \cite{Nah} is utilized for training and testing our network. It contains 2103 paris of blurry and ground truth images in train dataset, and 1111 pairs in test dataset. Resolution of the image are 720p. The blurry image is generated by averaging a sequence (7-15) of continuous sharp images. Sharp image in the middle of sequence is regarded as ground truth. GOPRO dataset is regarded as benchmark by many deblurring algorithms like \cite{Nah} and \cite{Kupyn}. Although GOPRO dataset is widely used, it only employs noise-free images. For natural images, however, noise always accompanies with blur. To test our model on more real images, we add Gaussian noise with $variance = 0.001$ to original GOPRO\_Large dataset and create a new GOPRO-noise dataset with 1111 image pairs. A synthetic dataset in \cite{Kupyn} is adopted for training. Same as combination version of DeblurGAN in \cite{Kupyn}, we use both GOPRO train dataset and synthetic dataset to train our network.
\subsection{Training Process}
The proposed network is trained on NVIDIA GeForce GTX 1080 Ti GPU and tested on Mac Pro with 2.7 GHz Intel Core i5 CPU. Similar to \cite{Kupyn}, the input training pair is randomly croped as size of $256 \times 256$ after downsampled by a factor of two. Weights are initialized to follow Gaussian distribution with zeros mean and standard deviation $0.02$. For each iteration of optimization, 1 step is performed on discriminator $D$, followed by 2 steps on generator $G$ to prevent discriminator loss $\mathscr{L}_d$ from zero. The model is trained for 15 epochs within 2 days, comparing with 200 epochs for 6 days in \cite{Kupyn} . Furthermore, despite of instability GAN's training, our method converges to similar result for each and every training task, which demonstrates the robustness of our GAN architecture. 
\subsection{Result and Comparison}
Our test results are mainly compared with state-of-art GAN based deblurring network DeblurGAN \cite{Kupyn}. DeblurGAN defeats deep learning networks \cite{Sun} and \cite{Nah} on GOPRO dataset. Since the author posted the code online\footnote{https://github.com/KupynOrest/DeblurGAN}, we compare our network with DeblurGAN by directly adopting its uploaded network and latest trained weights. We test our model on GOPRO and GOPRO-noise test datasets.

Fig. \ref{Fig:03} illustrates the deblurred results of \cite{Kupyn} and our model. Blurry image in the first row is picked in GOPRO-noise dataset and the blurry one of second row is real natural image with motion blur taken by camera. According to local patches, although \cite{Kupyn} can deal with blur but its results suffer from grid artifacts, while our model with dark channel loss achieves sharper images without grid artifacts. Furthermore, for motion blurry image (second row), the sharp part in input image remains unchanged in our deblurred result, but extra grid artifacts are added to result of \cite{Kupyn}.

The quantitative performance of the proposed network on two dataset GOPRO and GOPRO-noise  is shown in Tab. \ref{tab:Average}. In our experiment, the coefficient of dark channel loss $\lambda_2 = 250 $($dc250$). The results are compared with same network without dark channel loss $dc0$, same network with extra perpetual loss $dc250p$ as well as DeblurGAN \cite{Kupyn}. All test images are downsampled by factor of two. The perpetual loss follows what it is in \cite{Kohler}. The proposed model performs best among the comparisons for both noise-free and noisy dataset. DeblurGAN performs less well owing to its grid artifacts. Perceptual loss leads to a worse result. Since GAN is good at preserving perceptual feature already, perceptual loss brings no extra constraints for the network. Comparison with dc=0 demonstrates that dark channel loss contributes to better result.

\begin{table}[!t]
	\caption{Table. 1 Average PSNR and SSIM. }
	\label{tab:Average}
	\begin{center}       
		\begin{tabular}{|l|l|llll|} 
			\hline
			Dataset  &  Metrics  &   \cite{Kupyn} & $dc0$ & $dc250$ & $dc250p$ \\
			\hline 
			Original    &  PSNR  &  26.63 & 26.70 & {\bf 27.01 }& 26.45 \\
			~  &  SSIM  &  0.8701  &  0.8798& {\bf 0.8813 }& 0.8680 \\
			\hline
			Noisy   &  PSNR   &   26.32& 26.53 & {\bf 26.83} & 26.31 \\
			~  &  SSIM   &   0.8524 & 0.8697 & {\bf 0.8707} & 0.8604 \\
			\hline
		\end{tabular}
	\end{center}
\end{table} 

\section{Conclusion}
To address deblurring problem using a CGAN based architecture and to tackle the issue with grid artifacts in GAN based deblurring methods, this paper incorporates a dark channel prior. The dark channel prior is employed by $L_2$ norm rather than $L_0$ in order to make it more friendly for network training. To validate the deblurring result on more nature images, a noise involved dataset is proposed. The proposed network shows a great deblurring performance for both synthetic and real blurry images. 




\end{document}